\documentclass[letterpaper, 10 pt, conference]{ieeeconf}  %

\usepackage{graphicx} %
\usepackage{xcolor} %
\usepackage{multirow}
\usepackage{tabularray}
\usepackage{cite}
\usepackage{array}
\usepackage{amsmath} %

\IEEEoverridecommandlockouts
\begin{document}

\title{Testing and Evaluation of Underwater Vehicle Using Hardware-In-The-Loop Simulation with HoloOcean}
\author{Braden Meyers and Joshua G. Mangelson
    \thanks{This work was funded under Department of Navy awards N00014-24-1-2301 and N00014-24-1-2503 issued by the Office of Naval Research. Parts of this work were also funded by Naval Sea Systems Command (NAVSEA), Naval Surface Warfare Center - Panama City Division (NSWC-PCD) and Naval Undersea Warfare Center - Keyport Division (NUWC-KPT) under the Naval Engineering Education Consortium (NEEC) Grant Program under award numbers N00174-23-1-0005 and N00178-23-1-0006.}%
  \thanks{B.~Meyers and J.~Mangelson are at Brigham Young University. They can be reached at: \texttt{\{bjm255, mangelson\}@byu.edu}. }
}
\maketitle

\begin{abstract}
Testing marine robotics systems in controlled environments before field tests is challenging, especially when acoustic-based sensors and control surfaces only function properly underwater. Deploying robots in indoor tanks and pools often faces space constraints that complicate testing of control, navigation, and perception algorithms at scale. Recent developments of high-fidelity underwater simulation tools have the potential to address these problems. We demonstrate the utility of the recently released HoloOcean 2.0 simulator with improved dynamics for torpedo AUV vehicles and a new ROS 2 interface. We have successfully demonstrated a Hardware-in-the-Loop (HIL) and Software-in-the-Loop (SIL) setup for testing and evaluating a CougUV torpedo autonomous underwater vehicle (AUV) that was built and developed in our lab. With this HIL and SIL setup, simulations are run in HoloOcean using a ROS 2 bridge such that simulated sensor data is sent to the CougUV (mimicking sensor drivers) and control surface commands are sent back to the simulation, where vehicle dynamics and sensor data are calculated. We compare our simulated results to real-world field trial results.
\end{abstract}

\section{Introduction}
\label{sec:intro}

Developing and deploying underwater robotic systems presents unique challenges, including high operational costs, safety risks, and extremely limited communication. 
During the development phase of hardware and software, these challenges make field testing particularly risky, especially when monitoring underwater autonomous behavior in real-time is not feasible. Many autonomy and control algorithms rely on feedback from onboard sensors, yet testing these systems before deployment is difficult without access to realistic simulation environments. As a result, high-fidelity simulation tools are essential for validating software and hardware in closed-loop scenarios prior to field trials.

In the Field Robotics Systems (FRoSt) Lab at BYU, we leverage the open-source HoloOcean simulator to develop and evaluate the behavior of controllers, navigation, and localization algorithms prior to real-world deployment. As the fidelity of simulation environments improves, tools for testing and evaluating software and hardware become more valuable. HoloOcean supports a variety of sensors, vehicles, and features that significantly simplify underwater robotic perception, localization, and autonomy development \cite{potokarHoloOcean2022, potokarHoloOceanSonar2022, potokarHoloOcean2024}. In this paper, we present several upgrades to the HoloOcean system that enable Hardware-in-the-Loop (HIL) and Software-in-the-Loop (SIL) testing of real-world vehicles. 

\begin{figure}[t]
    \centering
    \includegraphics[width=\columnwidth]{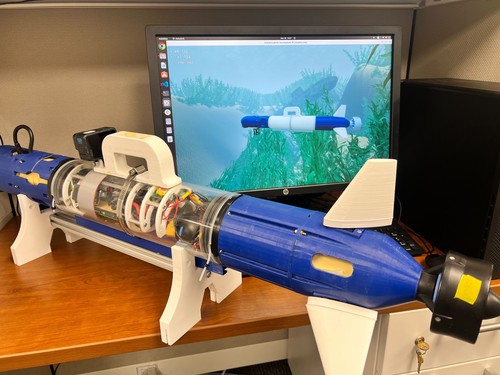}
    \caption{A CougUV robot is tested in the lab prior to field trials with a HIL setup with the HoloOcean simulator. Where simulated sensor data is sent to the CougUV and processed onboard. Control surfaces are actuated and commands are sent back to the simulator, where the vehicle's motion is calculated.}
    \label{fig:highlevel}
\end{figure}

The specific contributions of this paper include: 
\begin{enumerate}
    \item Integration of high-fidelity dynamics for torpedo autonomous underwater vehicles (AUVs) into the HoloOcean simulator; 
    \item Development of a ROS 2 interface for HoloOcean; and
    \item Demonstration and validation of HIL and SIL testing of a custom real-world AUV using these tools.
\end{enumerate}

Each of these new features has been made publicly available as part of the release of HoloOcean 2.0. Additional features released in HoloOcean 2.0 (and documented elsewhere) 
include an upgrade to Unreal Engine 5.3, additional vehicles, improved rendering, environment modifications, and various other improvements and bug fixes.

The rest of this paper is outlined as follows. In Sec. \ref{sec:related_work}, we summarize related work. In Sec. \ref{sec:Dynamics}, we provide an overview of the implementation of high-fidelity torpedo AUV dynamics that have been integrated into HoloOcean 2.0. In Sec. \ref{sec:ROS2}, we describe the HoloOcean ROS 2 bridge. In Sec. \ref{sec:HIL}, we present and analyze several experiments that demonstrate the ability to utilize HoloOcean for HIL and SIL testing. Finally, we conclude in Sec. \ref{sec:conclusion}.

\section{Related Work}
\label{sec:related_work}

Robotics simulators have advanced significantly in recent years, particularly in efforts to bridge the sim-to-real gap. Simulators that closely replicate real-world environments, vehicle dynamics, and sensor behavior provide a valuable platform for testing robotic systems before real-world deployment. In underwater robotics, several simulators have emerged that reduce the sim-to-real gap in different ways.

\subsection{HoloOcean}

HoloOcean \cite{potokarHoloOcean2024} (originally released in 2022 \cite{potokarHoloOcean2022}) leverages the Unreal tool chain to enable high-fidelity marine robotic perception and autonomy simulation with advanced sonar simulation support \cite{potokarHoloOceanSonar2022}. In its initial form, HoloOcean relied upon the built-in medium-fidelity physics engine of Unreal Engine for dynamics simulation, which is typically optimized for real-time performance in gaming applications. Despite this, HoloOcean supports realistic sensor placement and configuration to closely replicate real-world systems and supports multi-agent setups within complex underwater environments. However, HoloOcean is often cited for lacking ROS support and high-fidelity underwater dynamics.

\subsection{Marine Dynamics Simulation}

The Marine Systems Simulator (MSS) \cite{marinesystemssimulator}, originally released in 2004, is a MATLAB/Simulink toolbox for modeling and controlling marine craft, based on the equations from Fossen's \emph{Handbook of Marine Craft Hydrodynamics and Motion Control} \cite{fossen}. The Python Vehicle Simulator \cite{pythonvehiclesimulator} re-implements many of these models in Python, supporting underwater and surface vehicles, as well as ships. While Fossen's models are widely accepted as the standard for AUV and ASV dynamics, the simulators cited above are typically used as standalone packages for simulating vehicle dynamics and controls, but lack visualization and sensor simulation features. In this paper, we present an effort to integrate these capabilities with HoloOcean.

\subsection{HIL and SIL Testing and ROS Integration}

The Robot Operating System (ROS) \cite{ROS2} has enabled modular software architectures for robotics, and its integration with simulators has facilitated HIL and SIL testing. Gazebo \cite{gazebo}, one of the most widely used simulators in robotics research, supports ROS out-of-the-box and is commonly used for HIL and SIL testing. However, Gazebo is limited in the realism of its simulation environments and in its perceptual sensing rendering capabilities.   
In this paper, we also present on an effort to develop a bridge connecting HoloOcean with ROS 2. 

A recent review of underwater simulators for digital twinning and  %
HIL/SIL simulation \cite{DigitalTwin} identified UWSim \cite{UWSIM}, Stonefish \cite{stonefish}, and UUV Simulator \cite{UUVSim} as the top tools for these applications. These simulators provide interfaces for integrating real-time control systems and hardware with simulated environments. OceanSim \cite{song2025oceansim} is a recently released underwater simulator focused on underwater vision-based sensing also supports ROS integration, but, as far as we know, has yet to be used for SIL/HIL testing.

\vspace{0.3cm}
\section{High-Fidelity Dynamics in HoloOcean}
With the release of HoloOcean 2.0.1, the interface for leveraging high-fidelity custom dynamics within HoloOcean has been given a significant upgrade. HoloOcean provides multiple control schemes for managing the dynamics of an agent or multiple agents. The custom dynamics control scheme exposes implementation of the dynamics to the user, thus enabling full configurability. In this paper, we present an effort to integrate these tools with the Fossen dynamics models \cite{fossen} for an underwater torpedo-like AUV with arbitrary fin placement.

In the following subsections, we first provide an overview of the new architecture used to implement custom dynamics in HoloOcean 2.0.1. We then outline the high-level equations of motion used to simulate torpedo vehicle dynamics. Finally, we outline the dynamics equations that enable the application of these models to AUVs with arbitrary fin configurations.

\subsection{Hydrodynamics Implementation Architecture}
The new custom dynamics implementation architecture for HoloOcean consists of two main components: (1) a vehicle dynamics model that implements the hydrodynamic equations of motion for each agent, and (2) a simulation interface manager that connects the dynamics model to the HoloOcean environment. This structure allows different types of vehicles and control modes to be supported consistently across simulation scenarios.

The vehicle controller for each agent can be configured with custom parameters during scenario configuration to match a specific vehicle and is equipped with built-in depth and heading control or manual control of the control surfaces. 
The default model parameters are for the REMUS 100 \cite{HII_REMUS100}.
A summary of the configurable parameters for each vehicle is given in Table \ref{tab:fossen-params}. All parameters are presented and explained in the Fossen's Handbook \cite{fossen}. 

Currently, only the torpedo AUV vehicle dynamics are available out-of-the-box in HoloOcean 2.0.1 for users, but vehicle dynamic models could be added to this framework easily using the same code structure as provided in Fossen's Python Vehicle Simulator \cite{pythonvehiclesimulator}.

\subsection{Torpedo Vehicle Dynamics}
\label{sec:Dynamics}

\begin{table}[t]
    \centering
    \renewcommand{\arraystretch}{1.3}
    \begin{tabular}{| >{\centering\arraybackslash}m{2cm} | >{\centering\arraybackslash}m{5.5cm} |}
        \hline
        \textbf{Category} & \textbf{Parameters} \\
        \hline
        Environmental & Water Density, Gravity, Currents \\
        \hline
        Physical & Mass, Length, Diameter, Inertia \\
        \hline
        Hydrodynamic  & Low-speed Linear Damping \\
        \hline
        Hydrostatic  & Center of Bouyancy and Mass locations \\
        \hline
        Control Surfaces & Time Constants, Lift/Thrust Coefficients, Fin Positions, Fin Area \\
        \hline
        Autopilot & Pitch-Depth PID Gains, Heading SMC Gains \\
        \hline
    \end{tabular}
    \caption{List of configurable parameters for HoloOcean's torpedo vehicle dynamics.}
    \label{tab:fossen-params}
\end{table}

\begin{figure*}[t]
    \centering
    \includegraphics[width=\linewidth]{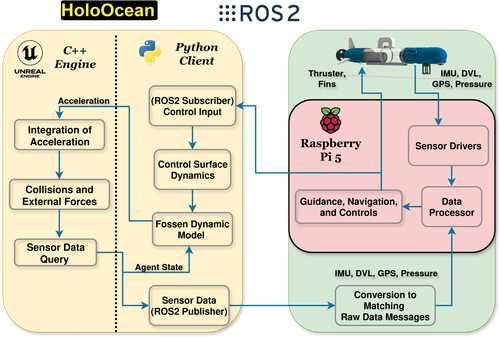}
    \caption{Software HIL and SIL flow diagram demonstrating the steps of vehicle dynamics calculations and implementation in the simulator. Sensor data output and control surface inputs are shown using the HoloOcean ROS 2 bridge in HoloOcean 2.0. Simulated sensor data can replace and mimic sensor driver output and control surface commands are transmitted to the vehicle actuators and to the simulation.}
    \label{fig:fossen}
\end{figure*}

High-fidelity dynamics models based on Thor Fossen's equations of motion for underwater vehicles \cite{fossen} are available open-source with the Python Vehicle Simulator \cite{pythonvehiclesimulator}. 
These models represent the current state-of-the-art for marine vehicle simulation, with accurate modeling of hydrostatic forces, dissipative forces, system inertia, and control surface dynamics and effects.

The equation of motion in the body frame, originally presented by Fossen \cite{fossen}, is as follows:
\begin{equation}
    M\dot{\nu_r}  + C(\nu_r)\nu_r + D(\nu_r)\nu_r + g(\eta) = \tau. \label{eq:motion}
\end{equation}
\begin{equation}
    M = M_{RB} + M_A
\end{equation}
\begin{equation}
    C(\nu_r) = C_{RB} (\nu_r) + C_A(\nu_r)
\end{equation}
\noindent Here, $M_{RB}$ represents the system inertia matrix for the rigid body and $M_{A}$ accounts for the added mass effects. The body-frame velocity of the vehicle is denoted by $\nu$, and $\nu_r$ represents the vehicle's velocity relative to the surrounding flow, also expressed in the body frame. The Coriolis and centripetal forces caused by the rigid body are represented by $C_{RB}(\nu_r)$, while $C_{A}(\nu_r)$ corresponds to the Coriolis forces arising from the added mass. The damping effects, both linear and quadratic, are encapsulated by $D(\nu_r)\nu_r$. The term $g(\eta)$ represents the hydrostatic forces, which originate from gravity and buoyancy, along with their associated moments and torques. Finally, $\tau$ denotes the external forces, such as those generated by the fins and thruster.

The vehicle pose, denoted by $\eta = [x, y, z, \phi, \theta, \psi]^T$, is a six-dimensional column vector. The orientation in terms of roll, pitch, and yaw angles is given by $\phi$, $\theta$, and $\psi$, respectively, and the linear and angular velocity vector in the body-fixed frame is $\nu = [u, v, w, p, q, r]^T$.

HoloOcean uses the equations of motion of the vehicle models to generate accelerations for each vehicle based on the state and control input at each simulation step. These accelerations are passed to the Unreal Engine 5 physics engine to handle collisions and other external forces. This flow is shown in the yellow box on the left side of Figure \ref{fig:fossen}.

\label{sec:DR}

\subsection{Torpedo Fin Dynamics}

Torpedo AUVs in HoloOcean are characterized by a single thruster and $N$ fins that can be actuated to induce pitch, roll, or yaw when the vehicle is in motion.
In an effort to support AUVs with an arbitrary number of fins, we have modified the dynamics models presented by Fossen as outlined in this section. 

The location of each fin is represented by the Center of Pressure (COP) with respect to the body frame of the vehicle, which we place at the Center of Mass (COM) for simplicity. The body-fixed frame defines the $\mathbf{x_b}$ axis along the longitudinal direction (forward), the $\mathbf{z_b}$ axis downwards, and the $\mathbf{y_b}$ axis to the starboard (right).

\begin{figure}[t]
    \centering
    \includegraphics[width=\linewidth]{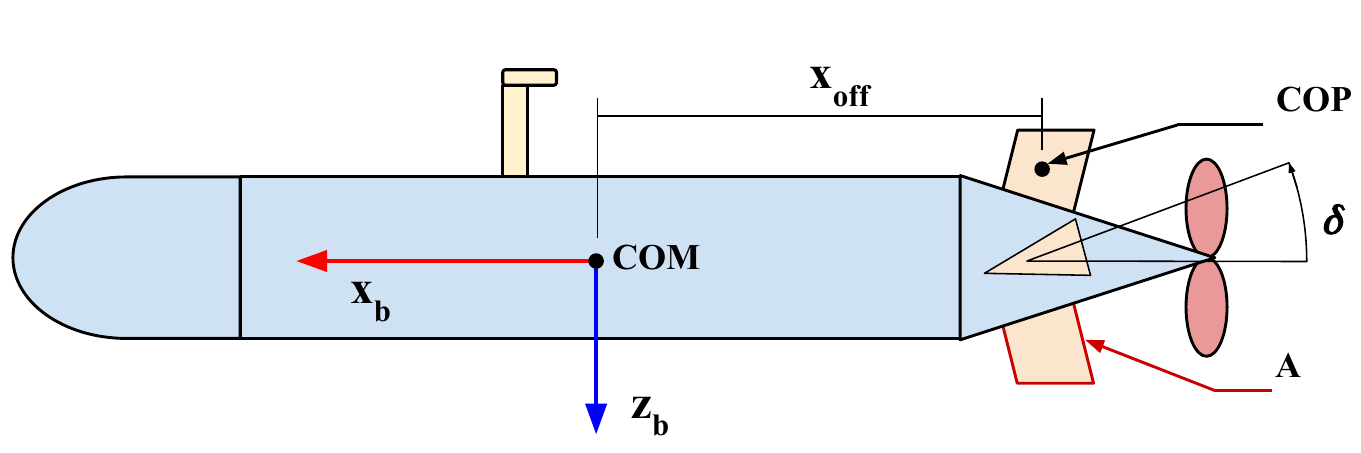}
    \caption{Side view of a Torpedo AUV fin diagram.}
    \label{fig:side_view}
\end{figure}

\begin{figure}[t]
    \centering
    \includegraphics[width=\linewidth]{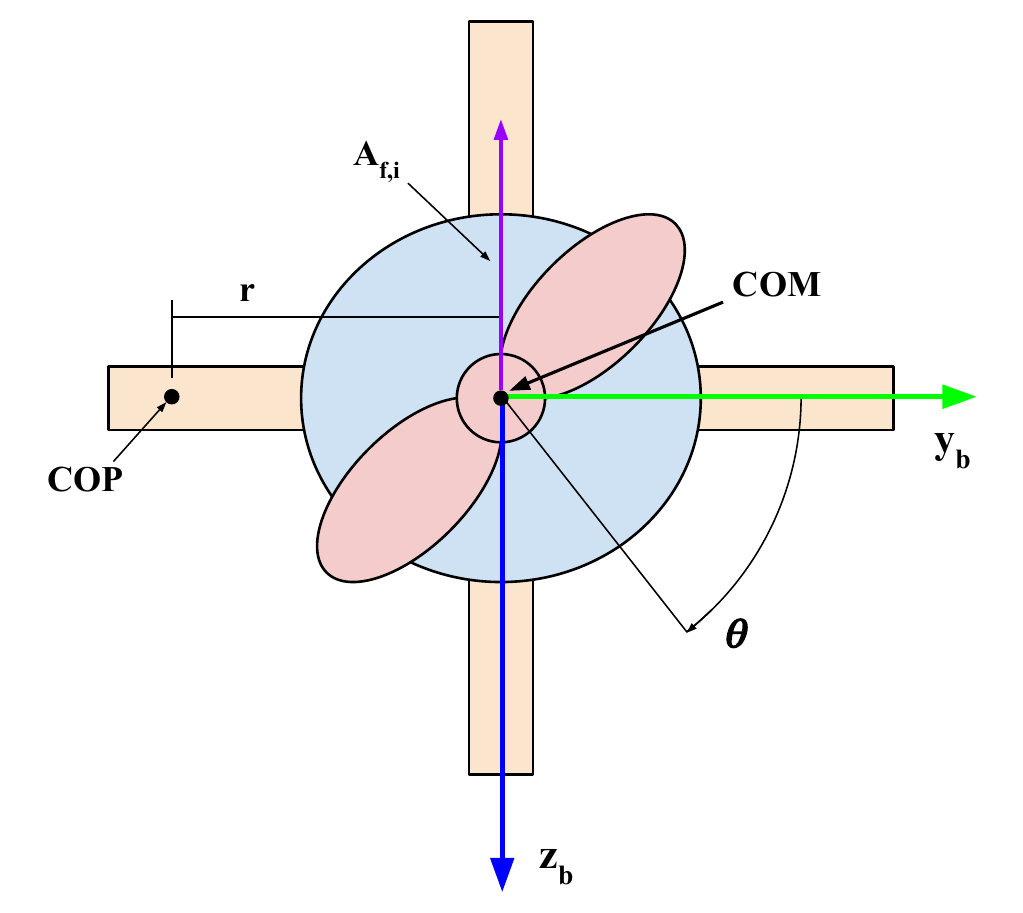}
    \caption{Back view of a Torpedo AUV fin diagram.}
    \label{fig:back_view}
\end{figure}

The offset of the COP of the fin from the COM is represented by three coordinates: $\mathbf{x_{off}}$, $\mathbf{r}$, and $\mathbf{\theta}$.
\begin{itemize}
    \item $\mathbf{x_{off}}$ represents the longitudinal distance from the COM to the COP along the $\mathbf{x_b}$ axis. This is depicted in the side view of the torpedo (Figure \ref{fig:side_view}), where the $\mathbf{x_{off}}$ dimension extends from the COM to the COP of the fin.
    \item The radial distance $\mathbf{r}$ defines how far the COP is from the $\mathbf{x_b}$ axis in the $\mathbf{y_b}-\mathbf{z_b}$ plane. As shown in Figure \ref{fig:back_view}, $\mathbf{r}$ is the perpendicular distance from the center (along the $\mathbf{x_b}$ axis) to the COP.
    \item The angle $\theta$ specifies the angular position of the COP around the $\mathbf{x_b}$ axis in the $\mathbf{y_b}-\mathbf{z_b}$ plane. As shown in Figure \ref{fig:back_view}, $\theta$ is measured counter-clockwise from the positive $\mathbf{y_b}$ axis to the projection of the COP onto the $\mathbf{y_b}-\mathbf{z_b}$ plane.
\end{itemize}
These three coordinates ($\mathbf{x_{off}}$, $\mathbf{r}$, $\theta$) precisely locate the COP of each fin with respect to the COM, allowing for a flexible representation of various fin configurations. This system enables users to accurately match the number of fins and their specific locations to real-world torpedo AUV designs. The term $\mathbf{A}$ in Figure \ref{fig:side_view} denotes the fin area outlined in red. Furthermore, the angle $\mathbf{\delta}$, shown in Figure \ref{fig:side_view}, represents a positive deflection angle of the control surface, which is used to calculate the forces and moments on the vehicle. A positive deflection is represented as the positive rotation around an axis $\mathbf{A_{f,i}}$, shown in figure \ref{fig:back_view} that starts at the $\mathbf{x_b}$ axis and increases along the $\mathbf{y_b}-\mathbf{z_b}$ plane through the point of rotation of the fin. 

To precisely model the hydrodynamic forces and moments exerted by each control fin on the vehicle, we define a force vector $\boldsymbol{\tau}_i$ for each individual fin $i \in \{1, \ldots, N\}$. This vector $\boldsymbol{\tau}_i = [F_{x,i}, F_{y,i}, F_{z,i}, M_{x,i}, M_{y,i}, M_{z,i}]^T$ represents the forces ($F_x, F_y, F_z$) and moments ($M_x, M_y, M_z$) acting on the body frame of the vehicle due to fin $i$. These forces and moments are calculated through the following sequence of steps:

\subsubsection{Relative Velocity Calculation} The relative velocity $\mathbf{v_{r,i}}$ for fin $i$ is defined as the component of the linear velocity of the vehicle relative to the fluid, specifically projected onto the plane of the fin. If $v_x$, $v_y$, and $v_z$ are the components of the vehicle linear velocity with respect to the body-fixed frame relative to the fluid, and $\theta_i$ is the angular position of fin $i$ around the $\mathbf{x_b}$ axis, then the effective relative velocity in the 
plane of the fin is given by:
\begin{equation}
\mathbf{v_{r,i}} = \sqrt{v_x^2 + (v_y \sin{\theta_i})^2 + (v_z \cos{\theta_i})^2} \label{eq:relative_vel}
\end{equation}
    This formulation accounts for the component of the flow velocity perpendicular to the span of the fin.

\subsubsection{Fin Hydrodynamic Force} The force $\mathbf{f_i}$ generated by fin $i$ is primarily a lift-like force perpendicular to the incident flow for small angle deflections of the fin. In particular, 
\begin{equation}
    \mathbf{f_i} = \frac{1}{2} \rho v_{r,i}^2 A_i C_{L,i}\delta_i, \label{eq:fin_force}
\end{equation}
    where $\rho$ is the water density, $A_i$ is the area of the fin $i$, $C_{L,i}$ is the lift coefficient, and $\mathbf{\delta_i}$ is the deflection of fin $i$.
    
\subsubsection{Fin Force Vector in Body Frame} Assuming the fin force $\mathbf{f_i}$ acts primarily in the $\mathbf{y_b}-\mathbf{z_b}$ plane, orthogonal to the longitudinal $\mathbf{x_b}$ axis, its components in the body frame are:
\begin{equation}
\vec{F}_i = [0, \mathbf{f_i} \sin{\theta_i}, -\mathbf{f_i} \cos{\theta_i}]^T. \label{eq:force_vec}
\end{equation}
    This decomposition aligns $\mathbf{f_i}$ with the angular position of the fin, $\theta_i$. The $\mathbf{x_b}$ component of the fin force is typically considered negligible for control surfaces primarily generating lift.

\subsubsection{Fin Center of Pressure Location} The location of the COP for fin $i$, $\vec{R}_i$, relative to the vehicle COM in body-fixed coordinates, is determined by its longitudinal offset $\mathbf{x_{off,i}}$, radial distance $\mathbf{r_i}$, and angular position $\theta_i$ as:
\begin{equation}
    \vec{R}_i = [\mathbf{x_{off,i}}, \mathbf{r_i} \cos{\theta_i}, \mathbf{r_i} \sin{\theta_i}]^T \label{eq:fin_location}
\end{equation}

\subsubsection{Moment Calculation} The moment $\vec{M}_i$ generated by fin $i$ about the vehicle COM is calculated as the cross product of the COP position vector and the force vector:
\begin{equation}
    \vec{M}_i = \vec{R}_i \times \vec{F}_i\label{eq:moment_calc}
\end{equation}

\subsubsection{Individual Fin Total Force and Moment} The complete force and moment vector $\boldsymbol{\tau}_i$ for fin $i$ in the body frame is then assembled:
\begin{equation}
    \boldsymbol{\tau}_i = [0, F_{y,i}, F_{z,i}, M_{x,i}, M_{y,i}, M_{z,i}]^T \label{eq:tau_i_components}
\end{equation}
    Here, $F_{y,i}$ and $F_{z,i}$ are the $y$ and $z$ components of $\vec{F}_i$, and $M_{x,i}$, $M_{y,i}$, $M_{z,i}$ are the $x$, $y$, and $z$ components of $\vec{M}_i$.

\subsubsection{Total Control Force and Moment} The total force and moment acting on the vehicle due to all $N$ fins, $\boldsymbol{\tau}$, is the vector sum of the individual fin contributions:
    $$\boldsymbol{\tau} = \sum_{i=1}^{N} \boldsymbol{\tau}_i \label{eq:total_tau}$$

\section{ROS 2 Interface}
\label{sec:ROS2}

ROS 2 is a common middleware for robotic systems that allows for a distributed and modular software architecture \cite{ROS2}. Nodes and groups of nodes can be tested independently, which allows for simple switching between simulation and real-world deployment.

With the release of HoloOcean 2.0, we added an interface for ROS 2, which offers access to more features and other robotics tools from the large active user base.
HoloOcean users can now use the ROS 2 bridge to publish sensor data and send commands to and from the simulator using standard ROS 2 topics, enabling integration with a wide range of tools, GUIs, and robotic systems built on the ROS 2 infrastructure.

The ROS 2 bridge has been configured to work with the new implementation of high-fidelity dynamics as described above. Vehicle commands (actuator positions, controller setpoints for speed, depth, and heading, etc.) can be sent to a torpedo agent through the Fossen dynamics interface. 
During simulation, sensor data from the agent is received by the HoloOcean Python client and translated into standard ROS 2 message types.

Several example nodes are provided that send commands and subscribe to sensor data to interface with the main HoloOcean node that runs the Unreal Engine Simulation. Multiple control modes are supported from these topics, including forces on the agent; control surface command; or desired depth, heading, and speed. 

Figure \ref{fig:fossen} shows an example of a connection between the simulation stack (left) and the robot software stack (right), established through ROS 2. In this example, the software stack for a CougUV (a custom AUV developed by the FRoStLab at BYU) sends control surface commands to the simulator and the hardware. Simulated sensor data flows into the software stack with the same topics and message types used by the sensor drivers.

To simplify the setup and installation for users, we provide Docker images that have been built with all necessary dependencies to run the simulation and ROS 2 interface.

\section{HIL and SIL Simulation with CougUV}
\label{sec:HIL}
 
These simulation improvements with HoloOcean enable a HIL and SIL simulation workflow using the ROS 2 architecture. 

We validated our approach using a CougUV, a torpedo AUV robot developed as a low-cost, modular sensor platform for multi-agent autonomy research. The system is made from commercial-off-the-shelf components. 
The CougUV software stack is built on the ROS 2 architecture with algorithms running in ROS 2 nodes. ROS 2 sensor drivers publish the data from the sensors on ROS 2 topics as seen in the top right side of Figure \ref{fig:fossen}. By matching sensor data, message types, and sensor locations between the real and virtual systems, we are able to test the full CougUV software stack using the HoloOcean simulator, excluding only the real-world sensors and drivers.

With the closed-loop feedback of the simulator, we can analyze and evaluate our software stack on the robot computer for controls, localization, and other algorithms. Sensor update rates match the real sensor update rates, which can help determine proper update rates for the other algorithms using sensor data. %

With HoloOcean we configured the parameters listed in Table \ref{tab:fossen-params}, the center of mass and buoyancy, and all time constants/actuator parameters based on the 3D model of the robot, relevant data sheets, and calculations with approximated vehicle geometry. 

External disturbances in simulation were not applied, and limited external disturbances affected the real-world tests. In future work, we hope to simulate and test in conditions with currents and waves.  

\begin{figure}[t]
    \centering
    \includegraphics[width=\linewidth]{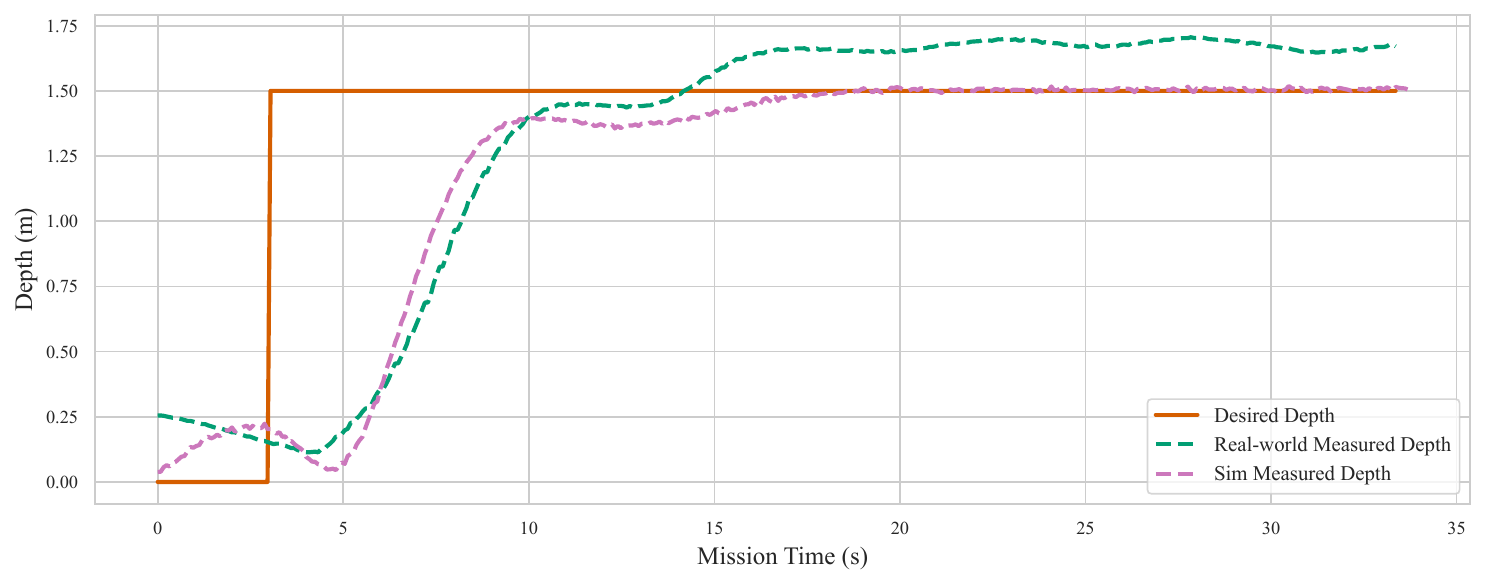}
    \caption{Comparison of depth controller response for the torpedo AUV in simulation and real world tests at a local reservoir.} 
    \label{fig:depth}
\end{figure}

\begin{figure}[t]
    \centering
    \includegraphics[width=\linewidth]{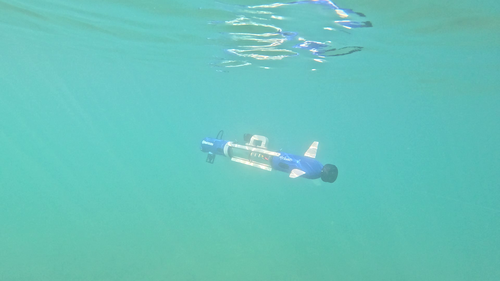}
    \caption{CougUV running a mission to hold a constant depth below the surface of the water. Field test performed at Bear Lake, UT in July 2025.}
    \label{fig:SideShot} 
\end{figure}

\subsection{Depth From Surface}
Maintaining a commanded depth from surface is a common autonomous behavior for AUV systems. We developed a nested pitch-depth controller that takes in desired depths and actuates the fins to pitch the vehicle and minimize the error between the desired and actual depth. 
Relative depth is calculated on the real system using a calibrated Blue Robotics pressure sensor and an estimate of the specific weight of the water. The depth sensor has a relative accuracy of ± 2 mbar (2 cm in fresh water). 
Pitch and pitch rate were measured with an onboard Inertial Measurement Unit (IMU).

After collecting data from the pressure sensor, we configured the simulated depth sensor in HoloOcean to publish at the same frequency (10 Hz) as our CougUV system, and a standard deviation of 0.01 m (1 cm). We performed over 50 simulation runs in HoloOcean to test and tune different proportional and derivative gains that achieved the desired rise time and damping for our system. During the simulation runs, we were able to verify that the fins responded with the correct rotation to perform the desired maneuvers. We then conducted real-world testing both in a pool and in a nearby reservoir with promising preliminary results. A comparison of the depth controller response in both sim and real-world for the same commanded mission is shown in Figure \ref{fig:depth}. An image of the vehicle conducting these real-world tests is shown in Figure \ref{fig:SideShot}.

\begin{figure}[t]
    \centering
    \includegraphics[width=\linewidth]{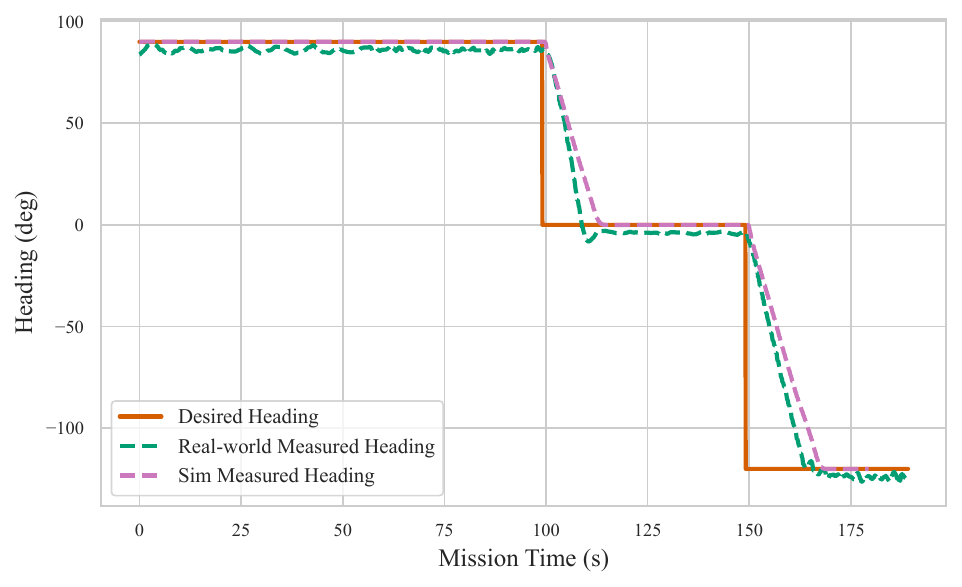}
    \caption{Comparison of heading controller response for the Torpedo AUV in simulation and the real world. Control parameters tuned in simulation transferred fairly well to real-world testing.}
    \label{fig:heading}
\end{figure}
\begin{figure}[b]
    \centering
    \includegraphics[width=\linewidth]{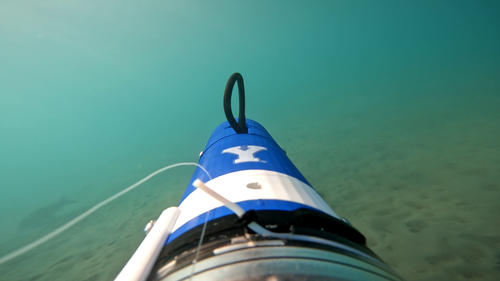}
    \caption{View from camera mounted on CougUV maintaining a constant altitude following the contours of the underwater terrain. Field test performed at Bear Lake, UT in July 2025.}
    \label{fig:DFBFish}
\end{figure}

\subsection{Heading}
Commanded heading for the CougUV can either be provided manually or via a high-level waypoint manager. Similar to previous tests, we ran simulations with heading step inputs to tune the controller. Heading of the vehicle is calculated with an onboard magnetometer that provides global heading when calibrated. 

Real-world results shown in Figure \ref{fig:heading} for the heading controller were less comparable to the simulated results especially for control when the vehicle was on the surface of the water. This is primarily because the top fin that controls the heading of the vehicle is not always completely submerged with the small waves in the water. These surface dynamics are not modeled in the simulation, but the response when the vehicle was completely underwater was more comparable to the simulated results. Future improvements to the simulator will account for water surface dynamics.

\begin{figure}[t]
    \centering
    \includegraphics[width=\linewidth]{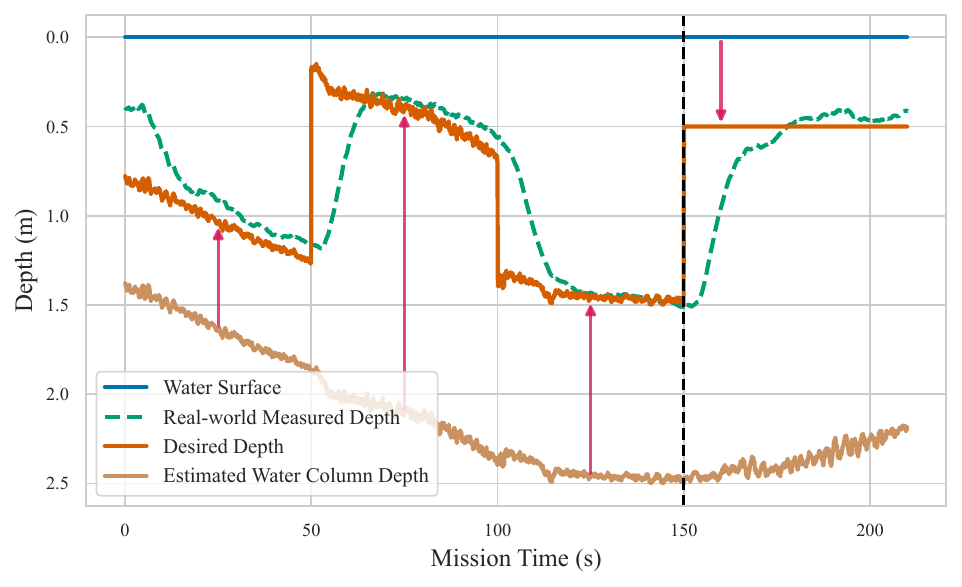}
    \caption{Depth Control real-world field test where modes of operation transitioned from a fixed altitude from the bottom for three different altitudes in succession to maintaining a depth-from-surface at 150 seconds from the start of the mission, where all depths are plotted relative to the water surface. The control parameters tuned in simulation directly translated to real-world use.}
    \label{fig:WaterColumn}
\end{figure}

\begin{figure}[t]
    \centering
    \includegraphics[width=\linewidth]{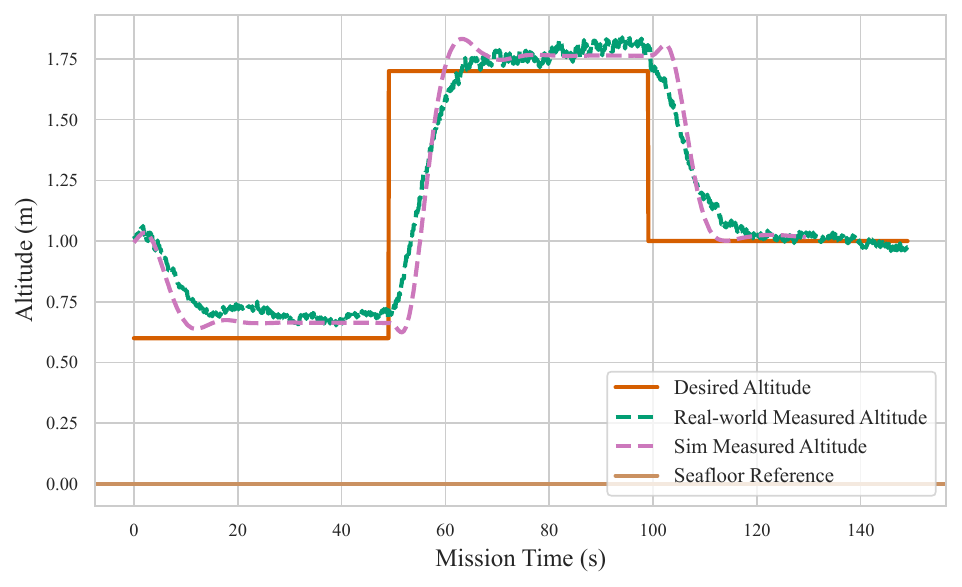}
    \caption{Altitude controller response for the Torpedo AUV in simulation compared to the actual real-world depth from Figure \ref{fig:WaterColumn} for the mission duration from 0 to 150 seconds, where altitude is plotted relative to the seafloor.}
    \label{fig:Altitude}
\end{figure}

\subsection{Altitude From Bottom}

Maintaining a fixed altitude is another common behavior for torpedo AUV systems. In this context, a vehicle follows the contours of underwater terrain, often to enable imaging of the seabed. This behavior is especially challenging to test and refine, as the vehicle often dives beyond the operator's line of sight, and a poorly-tuned controller can quickly lead to collisions with the bottom. 

Following successful results in the depth-from-surface missions, we applied the same control parameters in both SIL and HIL simulations to evaluate altitude-hold behavior. These simulations included models for vehicle dynamics, collisions, and altitude sensing to provide early-stage validation of the approach. We tested on two different terrain types in simulation, a flat bottom and a more aggressive terrain with bumps over 4 meters tall.  %
Because the vehicle is not equipped with a forward-facing sensor, a higher altitude was required to prevent collisions with the bottom.

Altitude was measured using a Doppler Velocity Log (DVL), model A50 from Waterlinked,
which provides both altitude and navigation information. Pitch and pitch rate were monitored using the vehicle's onboard IMU.

To validate and compare the controller in a real environment, we deployed the CougUV in a nearby reservoir where the seabed gradually sloped down away from shore, as shown in Figure \ref{fig:DFBFish}. The vehicle responded properly to step inputs in altitude and followed the terrain as shown in Figure \ref{fig:WaterColumn}. This demonstrates a strong correlation with the simulation results of the same mission on a flat terrain as shown in Figure \ref{fig:Altitude}. %

In most advanced applications, simulation tools could be used to estimate the minimum allowable altitude based on prior knowledge of the contours of the underwater terrain. Additionally, more sophisticated sensor models could evaluate performance in challenging conditions, such as DVL dropout or outlier altitude readings.

\subsection{Computational Load}

Running all the mission processes on a small, low-cost vehicle computer can strain system resources and potentially lead to unexpected behavior or mission failure. To monitor this, we track the Central Processing Unit (CPU) load and Random Access Memory (RAM) usage of the computer at 1 Hz during operation.  

To evaluate this, we performed both HIL and SIL simulations where the full mission software stack was run on the CougUV onboard computer and vehicle motion and sensor data were simulated in HoloOcean. %
The same methods for measuring CPU and RAM usage were used in simulation and in field trials.  

The data shown in table \ref{tab:cpu-ram-comparison} were taken from the same mission run in a HIL and SIL simulation and then in the nearby reservoir. Over the course of the mission CPU and RAM usages were approximately constant and very comparable. The CPU and RAM usage were consistently lower when compared in 5 different missions. This is likely due to the fact that fewer processes are running on the CougUV computer when running a simulated mission because the sensor drivers are run in simulation instead of onboard the vehicle computer. Depending on the sensors and sensor drivers, the effect of this discrepancy will vary. However, the SIL and HIL testing can be used to detect cases where the onboard processing requirement is certainly too high for real-time operation.

\begin{table}[h]
    \centering
    \renewcommand{\arraystretch}{1.3}
    \begin{tabular}{| >{\centering\arraybackslash}m{2cm} | >{\centering\arraybackslash}m{2.5cm} | >{\centering\arraybackslash}m{2.5cm} |}
        \hline
        \textbf{Mission Type} & \textbf{Avg. CPU Load (\%)} & \textbf{Avg. RAM Usage (\%)} \\
        \hline
        Real World & 27.4 & 8.29 \\
        \hline
        HIL Simulation & 26.5 & 8.21 \\
        \hline
    \end{tabular}
    \caption{CPU and RAM usage comparison between real world and simulated missions as a percentage of max CPU and RAM usage.}
    \label{tab:cpu-ram-comparison}
    \vspace{-1em}
\end{table}

\section{Conclusion}
\label{sec:conclusion}

In this paper, we document and present multiple improvements that form part of the HoloOcean 2.0 release. In particular, we describe the integration of high-fidelity dynamics for torpedo AUV vehicles and a ROS 2 interface that provide the tools necessary for HIL and SIL simulations. These capabilities aid the development of underwater robotic platforms and improve success rates in costly and high risk field trials. 

We demonstrate that real-time simulations can be used to ensure control surfaces and computational hardware are performing properly prior to field tests. We also show that software algorithms and autonomous behaviors can have comparable results in simulation and real-world tests when vehicle dynamics and sensor models are configured to match a real-world system.

We also demonstrate that torpedo AUV hydrodynamics can be modeled simply with parameters from 3D models, simple measurements, and data sheets. In future work, we plan to use system identification tools from ground truth measurements to estimate more accurate system parameters for the CougUV and compare real-world trajectories to the simulated trajectories.  This work can be easily extended to other vehicles with dynamic models that follow the same structure from Fossen's Python Vehicle Simulator \cite{pythonvehiclesimulator}.

Additional work is in progress on the HoloOcean simulator to improve the sim-to-real gap in marine environments. We continue to test more advanced autonomous behaviors and responses to environmental effects like currents and waves for both underwater and surface vessels. With HoloOcean, additional work can be done to develop and test real-time perception algorithms using the high-fidelity sonar sensors and ensure onboard vehicle computation can support the data processing. 

As robotic simulators continue to develop new tools, comparisons to real-world data are important to ensure the sim-to-real gap is decreasing. Validating underwater robotic simulators with real-world field trials is essential to simulation development. 

The code, Docker environment, and documentation used for this work are open source and can be found at https://github.com/byu-holoocean/holoocean-ros. Documentation for the HoloOcean simulator can be found at https://byu-holoocean.github.io/holoocean-docs.

\vspace{0.5cm}
\bibliographystyle{IEEEtran}
\bibliography{ref}
\end{document}